\pdfoutput=1
\documentclass[sigconf]{acmart}
\usepackage{bm}
\usepackage{multirow}
\usepackage{subcaption}
\usepackage{float}
\usepackage{multirow}

\usepackage{setspace}

\AtBeginDocument{%
  \providecommand\BibTeX{{%
    \normalfont B\kern-0.5em{\scshape i\kern-0.25em b}\kern-0.8em\TeX}}}

\copyrightyear{2020}
\acmYear{2020}
\setcopyright{acmcopyright}
\acmConference[MM '20]{Proceedings of the 28th ACM International Conference on Multimedia}{October 12--16, 2020}{Seattle, WA, USA}
\acmBooktitle{Proceedings of the 28th ACM International Conference on Multimedia (MM '20), October 12--16, 2020, Seattle, WA, USA}
\acmPrice{15.00}
\acmDOI{10.1145/3394171.3413931}
\acmISBN{978-1-4503-7988-5/20/10}

\begin{document}
\fancyhead{}

\title{Boosting Continuous Sign Language Recognition via Cross Modality Augmentation}

\author{
   Junfu Pu$^{1}$,
   Wengang Zhou$^{1,2}$,
   Hezhen Hu$^{1\scriptscriptstyle{*}}$,
   Houqiang Li$^{1,2}$
}
\affiliation{
   \institution{$^{1}$CAS Key Laboratory of Technology in GIPAS, EEIS Department, University of Science and Technology of China}
   \institution{$^{2}$Institute of Artificial Intelligence, Hefei Comprehensive National Science Center}
}
\email{pjh@mail.ustc.edu.cn, zhwg@ustc.edu.cn, alexhu@mail.ustc.edu.cn, lihq@ustc.edu.cn}



\renewcommand{\shortauthors}{Pu, et al.}

\begin{abstract}
Continuous sign language recognition~(SLR) deals with unaligned video-text pair and uses the word error rate~(WER), \emph{i.e.,} edit distance, as the main evaluation metric.
Since it is not differentiable, we usually instead optimize the learning model with the connectionist temporal classification~(CTC) objective loss, which maximizes the posterior probability over the sequential alignment.
Due to the optimization gap, the predicted sentence with the highest decoding probability may not be the best choice under the WER metric.
To tackle this issue, we propose a novel architecture with cross modality augmentation.
Specifically, we first augment cross-modal data by simulating the calculation procedure of WER, \emph{i.e.,} substitution, deletion and insertion on both text label and its corresponding video.
With these real and generated pseudo video-text pairs, we propose multiple loss terms to minimize the cross modality distance between the video and ground truth label, and make the network distinguish the difference between real and pseudo modalities.
The proposed framework can be easily extended to other existing CTC based continuous SLR architectures. 
Extensive experiments on two continuous SLR benchmarks, \emph{i.e.,} RWTH-PHOENIX-Weather and CSL, validate the effectiveness of our proposed method.
\end{abstract}




\maketitle

\begin{spacing}{0.85}
{\fontsize{8pt}{8pt} \selectfont
  \textbf{ACM Reference Format:}\\
  Junfu Pu, Wengang Zhou, Hezhen Hu, Houqiang Li. 2020. Boosting Continuous Sign Language Recognition via Cross Modality Augmentation. In \textit{Proceedings of the 28th ACM International Conference on Multimedia (MM ’20), October 12–16, 2020, Seattle, WA, USA.} 
  ACM, New York, NY, USA, 9 pages. 
  https://doi.org/10.1145/3394171.3413931}
\end{spacing}

\begin{figure}
\begin{center}
   \includegraphics[width=1.0\linewidth]{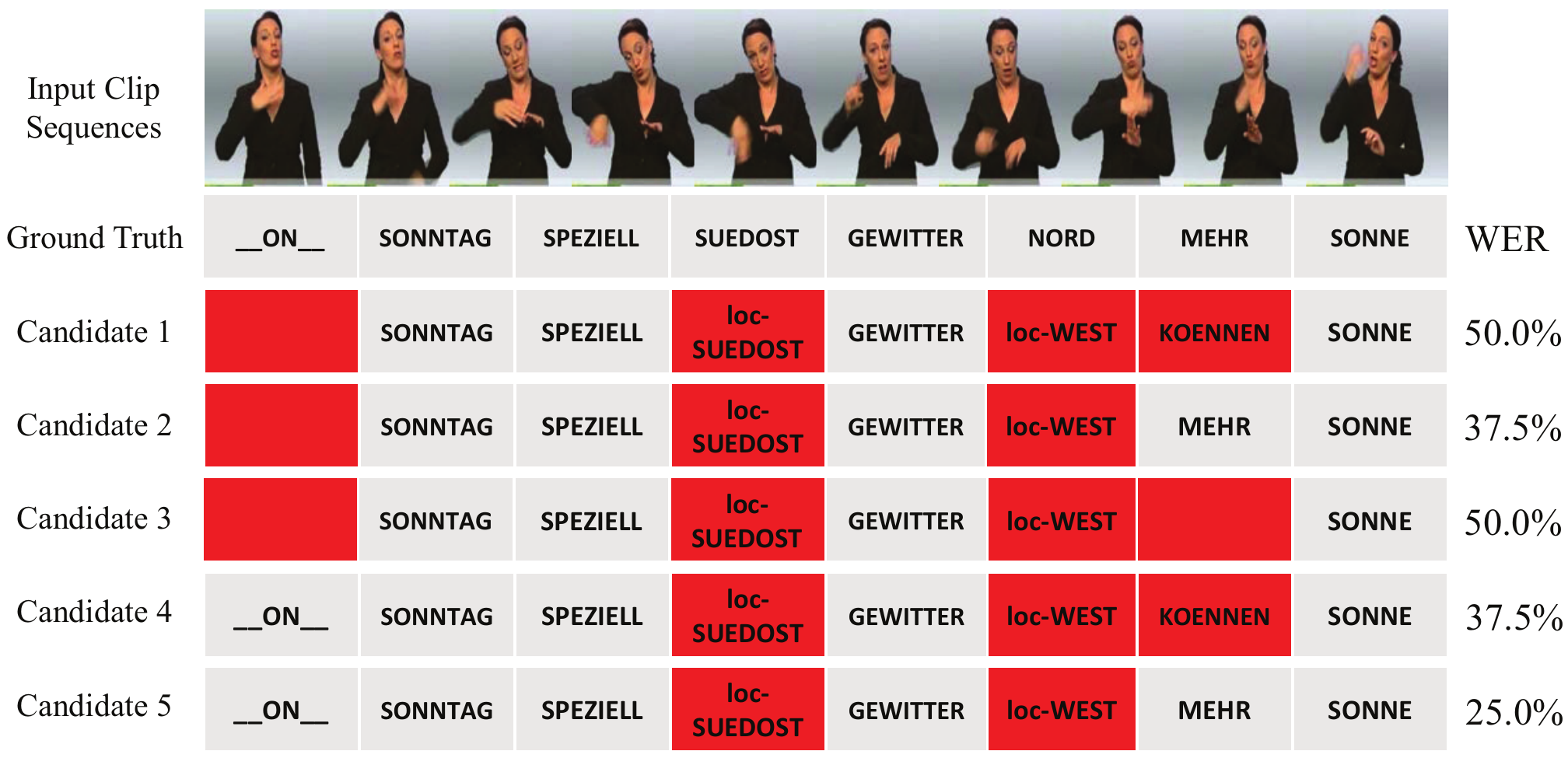}
\end{center}
  \caption{An example of decoding result. It shows five predicted sentences with descent decoding probability from Candidate 1 to 5. The box with red background denotes the false prediction.}
  \label{fig:intro}
  \Description{}
  \vspace{-0.5cm}
\end{figure}

\section{Introduction}
According to the official statistics from World Health Organization (WHO) in 2020, there are around 466 million people with disabling hearing loss, which accounts for over $5\%$ of the world's population.
Hearing loss leads to difficulty in hearing conversational speech.
As a result, people with hearing loss often use sign language for communication.
As a kind of visual language, sign language conveys semantic meanings by gestures and hand movements, together with facial expressions.
During the long-term evolution, sign language develops its own characteristic rules and grammar.
To facilitate such communication, many research efforts have been devoted to continuous sign language recognition (SLR), which aims to automatically identify the corresponding sign word sequence from a given sign video.
It's a transdisciplinary research topic which involves computer vision, natural language processing, and multimedia analysis, \emph{etc}.
Due to the expensive labeling cost, the continuous sign videos are generally weakly labeled, which means there is no alignment annotation of text sign words to video frames in the sign video.

Early works \cite{starner1998real,zhang2016chinese} on continuous SLR rely on hand-crafted visual features and statistical sequential models, \emph{i.e.,} Hidden Markov Model (HMM).
Recently, with the success of convolutional neural networks (CNNs) and recurrent neural networks (RNNs) in various computer vision tasks, more and more deep learning based sign language recognition algorithms have been proposed and achieved remarkable performance.
Among them, most of the state-of-the-art continuous SLR approaches \cite{cui2019deep,pu2019iterative,camgoz2017subunets,zhou2020spatial} utilize connectionist temporal classification (CTC), which is a popular technique to deal with sequence-to-sequence transformation without accurate alignment.
In CTC based methods, beam search algorithms are used for decoding, which iteratively produces word candidates.
Basically, the decoding precision and speed depend on the beam width.
The number of candidates for the decoded sequences is also equal to the beam width.
In practice, we choose the candidate with maximum decoding probability as the final predicted sentence.

However, due to the inconformity between CTC objective and evaluation metric, the candidate with maximum decoding probability may not be the best one under the evaluation metric, \emph{e.g.}, word error rate (WER).
For example, in Figure~\ref{fig:intro}, although Candidate 5 has the minimum decoding probability among all candidates, it is actually the best one with the lowest WER. 
To quantitatively illustrate this issue, we study the Top-$K$ WER on RWTH-PHOENIX-Weather based on the state-of-the-art method proposed in \cite{cui2019deep}.
Here, Top-$K$ WER is defined as follows: we choose the candidate with the lowest WER out of the $K$ decoded candidates and calculate the average WER over the whole dataset.
That is to say, Top-$K$ WER is a lower bound over the decoding results.
When the candidate with maximum decoding probability has the lowest WER for all testing samples, WER equals Top-$K$ WER.
According to our experiments, the WER on RWTH-PHOENIX-Weather testing set is $23.8\%$, while Top-5 WER decreases to $19.7\%$.
In order to bridge the performance gap, our target is to make the candidate with the best performance have the maximum decoding probability, which means to minimize the distance between such candidate and sign video.

\begin{figure}[ht]
\centering
    \begin{subfigure}[b]{\linewidth}
        \includegraphics[width=1.0\textwidth]{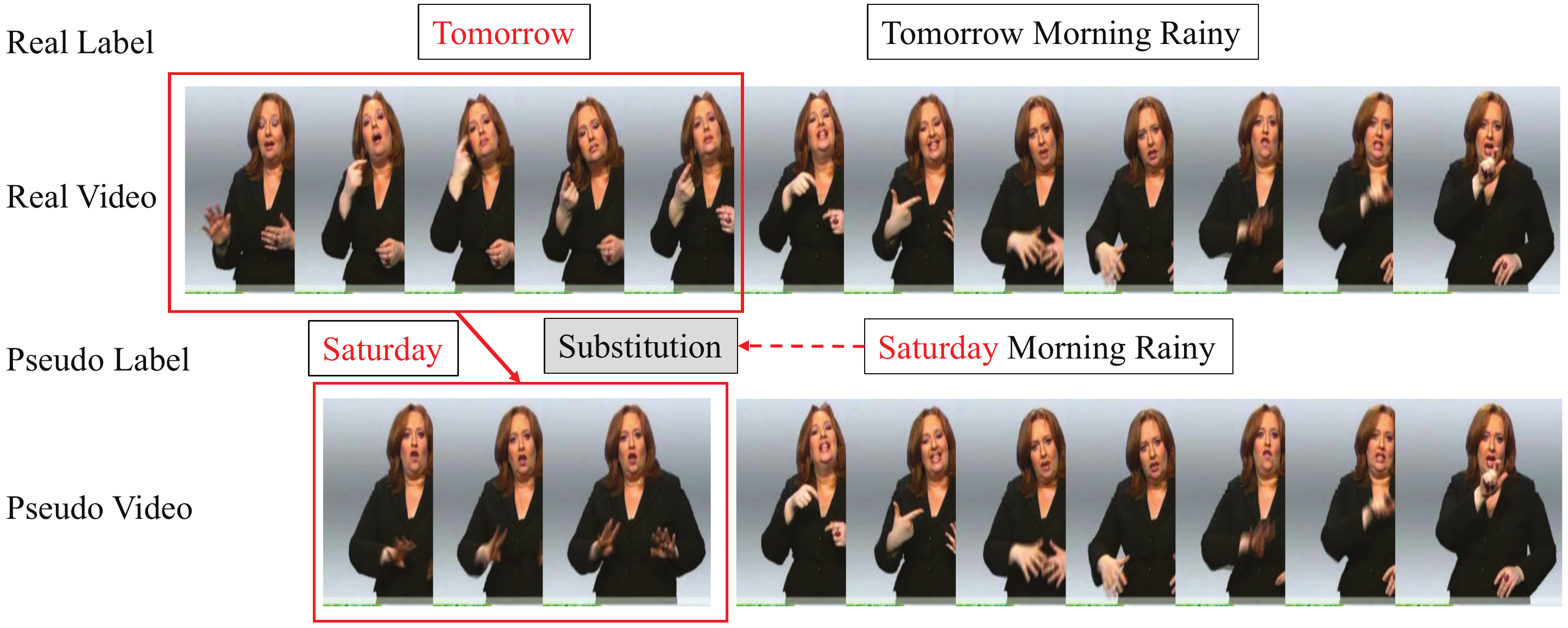}
        \caption{Illustration of the ``Substitution'' operation.}
        \label{subfig:edit_sub}
    \end{subfigure} %

    \begin{subfigure}[b]{\linewidth}    
        \includegraphics[width=1.0\textwidth]{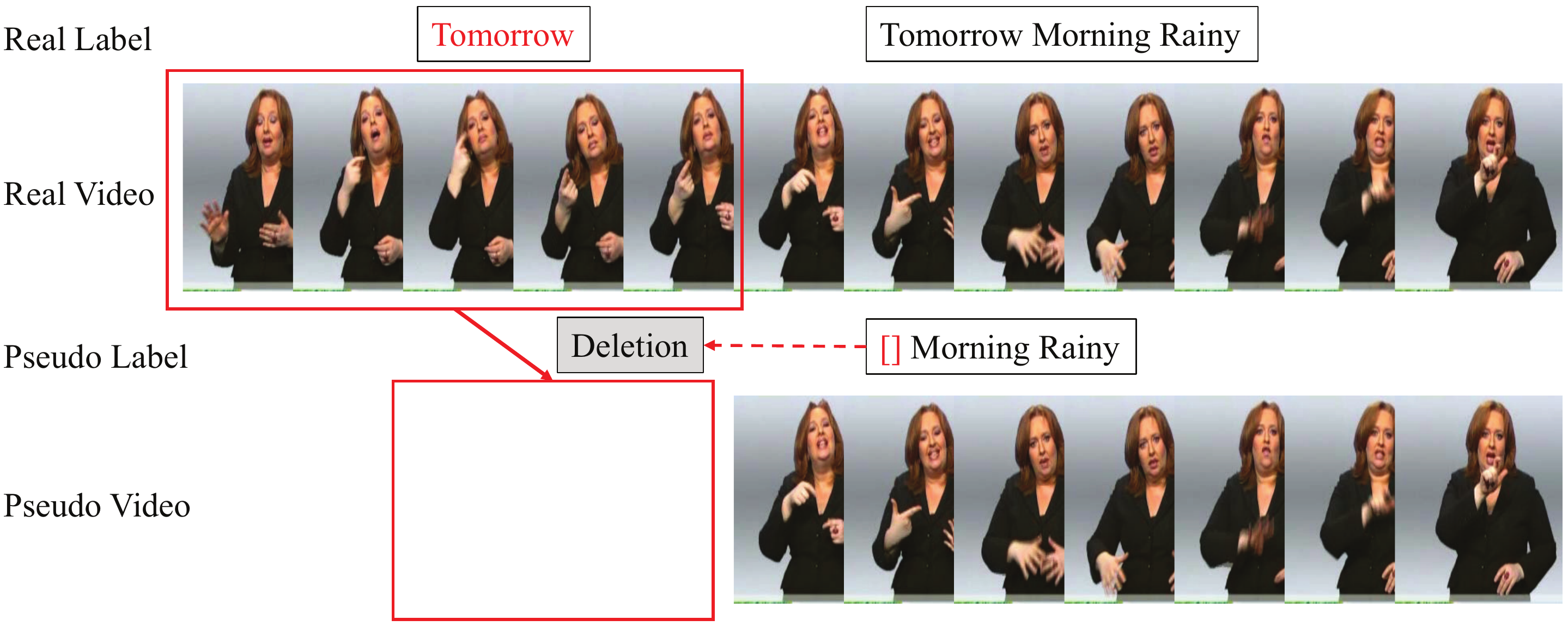}
        \caption{Illustration of the ``Deletion'' operation.}
        \label{subfig:edit_del}    
    \end{subfigure}

    \begin{subfigure}[b]{\linewidth}    
        \includegraphics[width=1.0\textwidth]{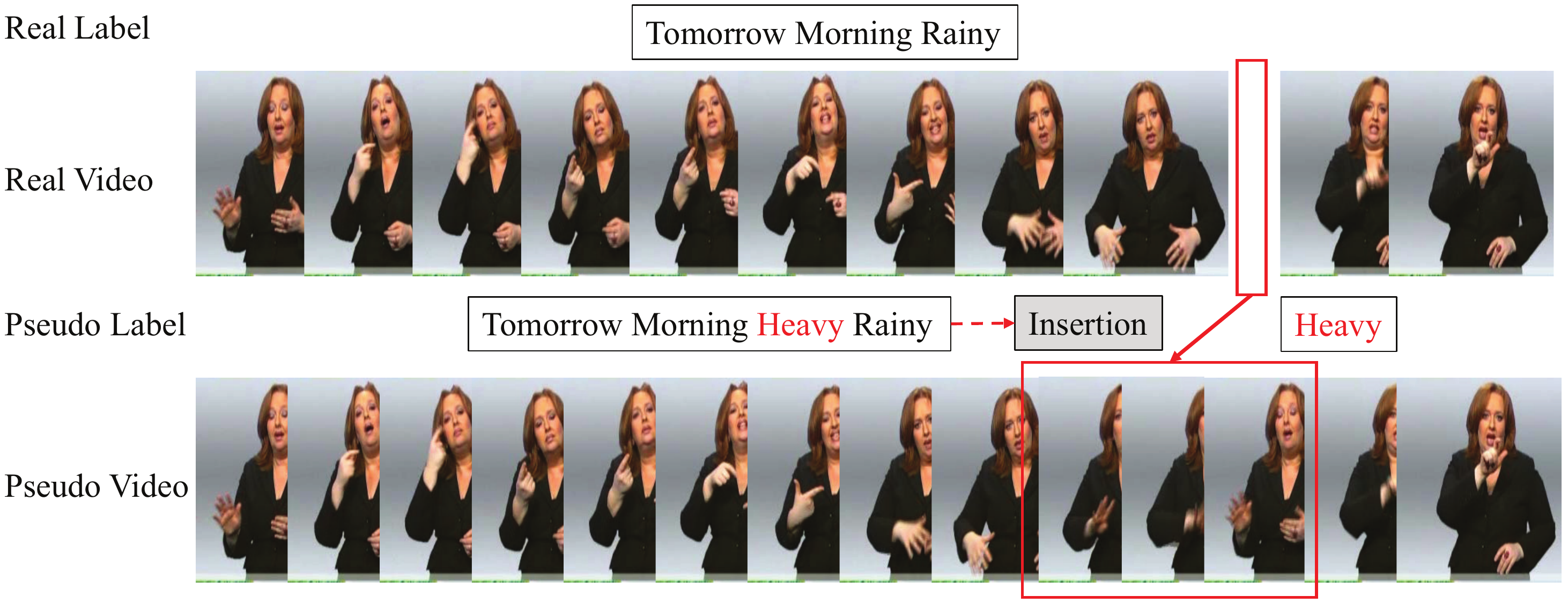}
        \caption{Illustration of the ``Insertion'' operation.}
        \label{subfig:edit_ins}    
    \end{subfigure} 
    \caption{Illustration of different kinds of editing operations.}
    \label{fig:editing_overview}
    \Description{}
    \vspace{-0.15cm}
\end{figure}

With the motivation discussed above, in this paper, we present a novel architecture for further boosting the performance of continuous SLR via cross modality augmentation.
In continuous SLR, WER is the most important evaluation metric, which is defined as the least operations, \emph{i.e.,} substitution, deletion, and insertion, to transform the target sequence to the reference sequence.
To simulate the calculation procedure of WER, we edit the sign video and corresponding text label following the same operations, as illustrated in Figure~\ref{fig:editing_overview}.
With such editing, we augment the cross modality data and obtain a pseudo video-text pair.
In order to minimize the cross modality distance from the video to ground truth label, while maximizing the distance from video to the pseudo text label, 
we propose a real-pseudo discriminative loss.
Besides, the objective includes alignment-based CTC loss for both real and pseudo video-text pair.
A cross modality semantic correspondence loss is also introduced to directly minimize the cross modality distance between real video and real ground truth text label.
The proposed framework can be easily extended to other existing CTC based continuous SLR architectures.
Extensive experiments on two continuous SLR benchmarks, \emph{i.e.,} RWTH-PHOENIX-Weather and CSL, demonstrate the effectiveness of our proposed method.

\begin{figure*}[ht!]
\begin{center}
   \includegraphics[width=0.97\linewidth]{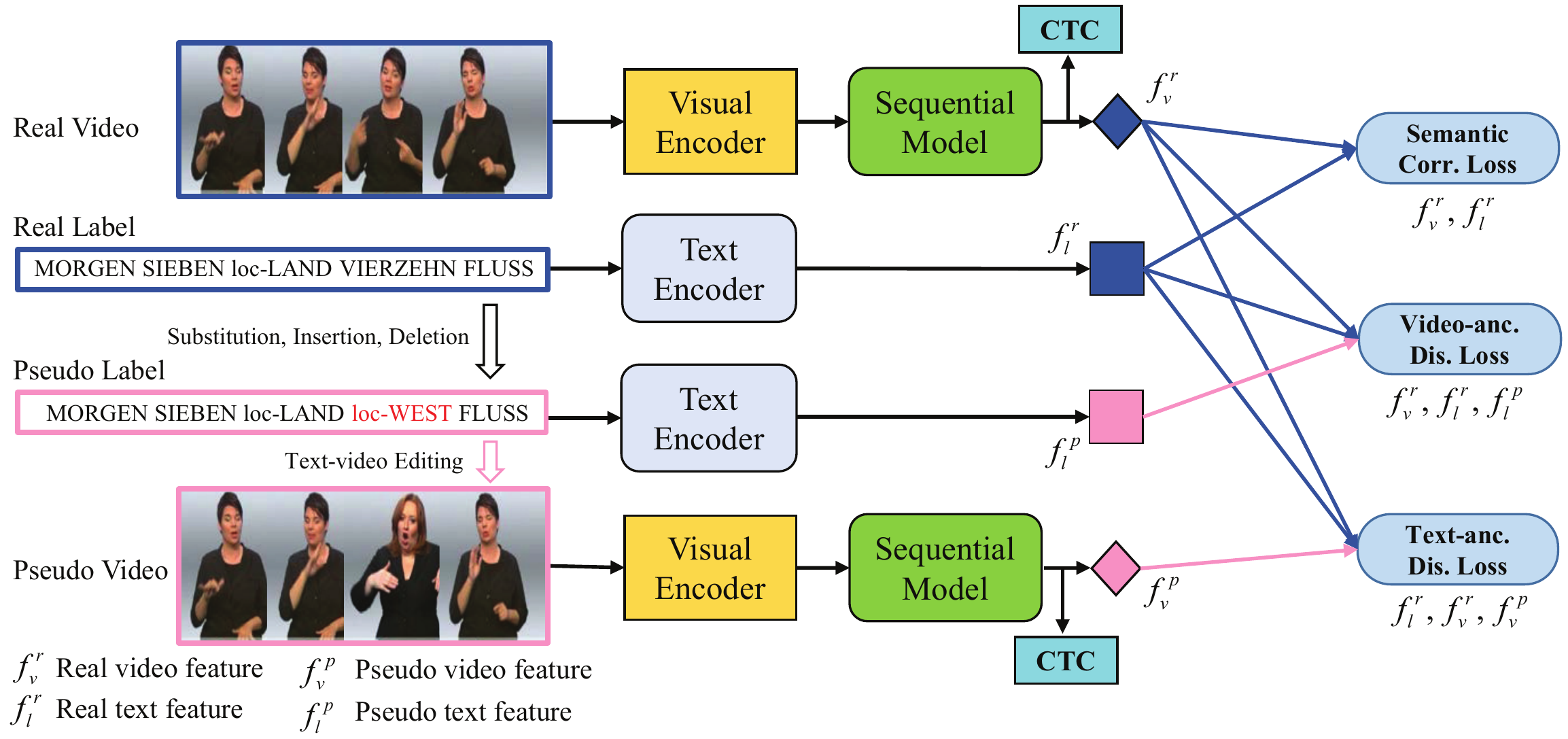}
\end{center}
  \caption{Overview of our proposed framework. The framework consists of a common CNN-TCN visual encoder, sequential model and text encoder. With the cross modality augmentation, we design multiple loss terms to optimize the architecture.}
  \label{fig:overview_architecture}
  \Description{}
  \vspace{-0.02cm}
\end{figure*}

\vspace{-2px}
\section{Related Work}
In this work, we briefly review the key modules and techniques in continuous sign language recognition.
Continuous SLR targets at translating the input video into a sequence of glosses in a consistent order.
First, the visual encoder transforms the input video into a high-dimensional feature representation.
Then the sequential module tries to learn the mapping from this feature representation to the corresponding text sequence.
To further refine the recognition result, iterative refinement strategy has been explored with promising results.
Besides discussing the above content, we also introduce existing data augmentation techniques in deep learning.

\textbf{Video representation learning.}
Discriminative feature representation is crucial for sign language recognition.
Early works concentrate on the hand-crafted features, such as HOG or HOG-3D~\cite{buehler2009learning, koller2015continuous}, motion trajectories~\cite{koller2015continuous, pfister2013large, evangelidis2014continuous} and SIFT~\cite{pfister2013large}.
These features are utilized for describing hand shapes, orientations or motion status.
With the advance of convolutional neural networks~(CNNs), many networks are designed for video representation learning.
They are based on 2D CNNs~\cite{wang2016temporal, he2016deep, simonyan2014two}, 3D-CNNs~\cite{tran2015learning, qiu2017learning, carreira2017quo, qiu2019learning} or a mixture of them~\cite{jiang2019sparse, zolfaghari2018eco, xie2018rethinking}.
For the task of continuous SLR, various CNNs have been investigated.
Oscar~\emph{et al.}~\cite{koller2017re} utilize GoogLeNet~\cite{szegedy2015going} as the visual encoder in an end-to-end iterative learning framework.
Pu~\emph{et al.} \cite{pu2018dilated} and Zhou~\emph{et al.} \cite{zhou2019dynamic} use 3D ResNet~\cite{qiu2017learning} and I3D~\cite{carreira2017quo} as the feature extractor to jointly model the spatial and temporal information, respectively.
There also exist methods~\cite{cui2017recurrent,cui2019deep} using 1D temporal CNNs after 2D CNNs to encode temporal dependency.
As one of them, DNF~\cite{cui2019deep} is becomes the most challenging competitor.

\textbf{Sequential learning.}
In continuous SLR, there are several popular sequential models, \emph{e.g.} hidden Markov model (HMM), recurrent neural network~(RNN) with connectionist temporal classification~(CTC) and encoder-decoder network, \emph{etc}.

HMM~\cite{wu2016deep, koller2017re, koller2018deep,koller2019weakly} is one of the most widely used sequential models.
Oscar \emph{et al.}~\cite{koller2019weakly} embed CNN-LSTM models in each HMM stream following the hybrid approach exploiting the state transitions and the sequential parallelism for sign language recognition.
The Recurrent Neural Networks (RNNs), \emph{e.g.} Long Short-Term Memory (LSTM)~\cite{hochreiter1997long}, Gated Recurrent Unit (GRU)~\cite{cho2014learning}, have been successfully applied to sequential problems, including speech recognition~\cite{graves2014towards}, video captioning~\cite{zhu2019attention, shi2019watch, yang2019structured}, machine translation~\cite{cho2014learning, sutskever2014sequence}, \emph{etc.}
In continuous SLR, bidirectional LSTM-CTC architecture~\cite{camgoz2017subunets,cui2017recurrent,pu2019iterative, cui2019deep} is employed as a basic model and becomes the most popular one.
Besides, there exist some works~\cite{huang2018video,guo2018hierarchical} adopting an attention-aware encoder-decoder network to learn the mapping between visual features and sign glosses.
Camgoz \emph{et al.}~\cite{cihan2018neural} also utilize the encoder-decoder architecture to extend sign language recognition to sign language translation.

\textbf{Iterative refinement.}
Given that the ground truth only provides sentence-level annotations without specific temporal boundaries for each gloss, continuous sign language can be treated as a weakly supervised problem.
Current frameworks usually contain a large number of layers and encounter the vanishing gradient problem for low layers, resulting in not fully optimized visual encoder.
Recent works demonstrate the importance of the alignments between video clips and sign glosses.
The video segment and sign gloss pairs can be treated as the trimmed video classification problem to enhance the visual encoder.
In this way, the whole architecture can be optimized in an iterative way for performance boosting.  
Cui \emph{et al.}~\cite{cui2019deep} and Zhou \emph{et al.} ~\cite{zhou2019dynamic} generate the pseudo glosses for video segments by aligning the output probability matrix and output glosses sequence through dynamic programming with notable performance gain.
Pu \emph{et al.} \cite{pu2019iterative} utilize utilize the soft Dynamic Time Warping (soft-DTW) as the alignment constraint with the warping path indicating the possible alignments between input video clips and sign words.

\textbf{Data augmentation in deep learning.} Data augmentation is a powerful method  to reduce overfitting, which can help the neural network to extract more information from the original dataset. 
Data augmentation encompasses a series of techniques enhance the quality and size of the training data.
It has been successfully applied in various deep learning based approaches.
There are many different data augmentation techniques in different tasks.
For image-based tasks, \emph{i.e.,} image classification, object detection, \emph{etc}, the image augmentation skills include geometric transformations (\emph{e.g.} rotation, flipping \emph{etc.}), color space transformation (\emph{e.g.} RGB to HSV), kernel filters, random erasing, \emph{etc.} 
For video-based tasks, \emph{i.e.,} action recognition, tracking, in addition to the image augmentation techniques, video augmentation is performed in temporal dimension by temporal random sampling.
In natural language processing tasks, \emph{e.g.} text classification, the operations to augment sentences include synonym replacement, random insertion, random swap, random deletion, \emph{etc.}
In existing techniques, the augmented sample share the same label with the original data, and the optimization loss keeps unchanged. In contrast, in our approach, the augmented video or gloss sentence no longer share the same semantic meaning with the original one. We take advantage of the fact and optimize the DNN model with novel triplet losses.

\section{Our Approach}
In this section, we first give an overview of our framework.
After that, we separately discuss the generation of pseudo video-text pairs, network architecture and loss design.

\subsection{Overview}
Our whole framework is illustrated in Figure~\ref{fig:overview_architecture}.
During training, given an input video and its corresponding text, we first perform the editing process to create the pseudo text.
At the same time, according to the same editing operations, we constitute the pseudo video based on the clip alignment, which is obtained from the refinement stage following~\cite{cui2019deep}.
Then we feed the real and pseudo video into the same recognition framework with shared parameters and calculate the CTC loss, respectively.
Further, we explore the relationship between real and pseudo video-text pairs by designing multiple losses to make the network aware of the editing operations and constrain the correspondence of cross-modal data.
The final optimization loss is a summation of the CTC loss, real-peseudo discriminative loss and cross modality semantic loss.
During the inference stage, the input video is fed into the backbone, \emph{i.e.,} the visual encoder, sequential model and CTC decoding model to output the final prediction text sequence.

\subsection{Pseudo Data Generation}
Following the definition of word error rate (WER), one of the most widely used evaluation metrics in continuous SLR, we generate pseudo video-text pair by editing raw video and label.
WER corresponds to the least operations of substitution, deletion and insertion to transform the target sequence into the reference sequence.
In order to simulate the calculation procedure of WER, the videos and labels are edited following these three operations, \emph{i.e.,}, substitution, deletion, and insertion.
Given a real video-text pair, we first substitute, insert or delete a word in the real text and repeat this operation a few times.
The inserted or substituted new word is randomly picked from the vocabulary in the training set.
On the other hand, we perform the same editing operations on the real video according to the alignment extracted in the refinement stage.
For a text label and its corresponding sign video, we perform $k$ editing operations, each of which is randomly taken from substitution, deletion, and insertion
The editing times $k$ is randomly sample from the range [1, $K$], where $K$ denotes the maximum editing operations. In this way, we obtain a pseudo video-text pair.

Figure~\ref{fig:editing_overview} illustrates three different editing operations.
Take ``Substitution'' as an example, as shown in Figure~\ref{subfig:edit_sub}, given a sign video with the label ``Tomorrow Morning Rainy'', we randomly replace a sign gloss.
In this case, sign gloss ``Tomorrow'' is replaced with ``Saturday''.
Thus, a new pseudo label sequence is generated with such editing operation.
After that, all frames corresponding to the sign word ``Tomorrow'' are also replaced by the frame segment with the meaning of ``Saturday'' from other videos.
Similarly, we can edit the video-text pair with another two operations, \emph{i.e.,}, insertion and deletion.
``Insertion'' indicates we randomly pick up a text word from the vocabulary and insert it into the original label sequence, while ``deletion'' means a sign word is randomly deleted from the label sequence.

\subsection{Network Architecture}
\textbf{Visual encoder.}
Visual encoder aims at encoding the input video into semantic feature representation.
It consists of a spatial encoder $E_{vs}$ followed by a temporal encoder $E_{vt}$ for spatial-temporal representation.
In our implementation, we use the same spatial-temporal backbone proposed in \cite{cui2019deep} considering its excellent performance.
GoogLeNet~\cite{szegedy2015going} is selected as our spatial encoder.
Temporal encoder contains the architecture of \emph{conv1d-maxpool-conv1d-maxpool}.
Specifically, the kernel sizes of 1D temporal convolutional layers and max pooling layers are set as 5 and 2, respectively.
The strides for all the layers in $V_t$ are set as 1.
Following these settings, the time length is reduced to a quarter of the original video with the receptive field as 16.
Given a sign video $\mathbf{V} = \{v_t\}_{t=1}^T$ with $T$ frames, the output, \emph{i.e.,} semantic feature representation, is defined as follows,
\begin{equation}\label{equ:cnn_tcn}
  \mathbf{\mathbf{F}} = E_{vt}(E_{vs}(\mathbf{V})),
\end{equation} 
where $\mathbf{V} \in {{\mathbb{R}}^{C_1 \times T\times H\times W}}$ and $\mathbf{F} \in {{\mathbb{R}}^{C_2 \times T/4}}$.

\textbf{Sequential model.}
The sequential model captures temporal dependency among the semantic feature representations generated by visual encoder, and learn the mapping between visual features and sign glosses.
We select Bidirectional Long Short-Term Memory~(BLSTM) $S_{bi}$, which captures the temporal dependency in both forward and backward time steps.
It takes the feature sequence as input and generates hidden states $\mathbf{F}_v$ as follows,
\begin{equation}
\label{equ:blstm}
  \mathbf{F}_v = S_{bi}(\mathbf{F}),
\end{equation} 
where $\mathbf{F}_v \in {{\mathbb{R}}^{C_3 \times T/4}}$ and $C_3$ indicates the units of the hidden states.
After that, the hidden states $\mathbf{H}$ is utilized as the input of a fully-connected layer $f_c$ and a softmax layer $f$ to generate the probability matrix for all the time steps as follows,
\begin{equation}
\label{equ:fc}
  \mathbf{\mathbf{P}} = f(f_c(\mathbf{F}_v)),
\end{equation} 
where $\mathbf{P} \in {{\mathbb{R}}^{N \times T/4}}$ and $N$ indicates the number of glosses.

\textbf{Text Encoder.}
For the semantic correspondence between the visual feature and gloss sequence, we utilize the text label encoder $E_T$ to map the gloss sequence $\bm s$ into the same latent space as the visual features as follows,
\begin{equation}
\label{equ:text encoder}
  \mathbf{F}_l = E_t(\bm{s}),
\end{equation} 
where $\mathbf{F}_l \in {{\mathbb{R}}^{C_3 \times T/4}}$ and a two-layer BLSTM is also utilized as the text encoder.

\subsection{Objective Function}
To optimize the network, we use three different kinds of loss functions, \emph{i.e.,}, alignment loss, real-pseudo discriminative loss, and cross modality semantic correspondence loss.
For each stream, in order to learn the alignment between video and text sequence, connectionist temporal classification (CTC) is introduced.
CTC is proposed to deal with two unsegmented sequences without accurate alignment.
CTC introduces a blank label out of the vocabulary to account for transitions and silence without precise meaning.
There may exist several alignment paths $\pi$ between the input sequence and target sequence.
The probability of each path $\pi$ is written as follows,
\begin{equation}\label{equ:ctc_path}
  p(\pi|\mathbf{V})=\prod_{t=1}^{T} p(\pi_t|\mathbf{V}),
\end{equation}
where $\pi_t$ is the label at time step $t$, $T$ is the number of frames in video.
A many-to-one mapping $\mathcal{B}$ is defined to remove reduplicated words and blank labels.
The conditional probability of the target sequence $\bm{s}$ is calculated as the summation of the probabilities of all alignment paths, which is formulated as follows,
\begin{equation}\label{equ:ctc_prob}
  p(\bm{s}|\bm{V}) = \sum_{\pi\in \mathcal{B}^{-1}(\bm{s})} p(\pi|\bm{V}),
\end{equation}
where $\mathcal{B}^{-1}$ is the inverse mapping of $\mathcal{B}$.
The final CTC objective function is defined as the negative log probability of $p(\bm{s}|\bm{V})$, written as follows,
\begin{equation}\label{equ:ctc_loss}
  \mathcal{L}_{\mathrm{CTC}} = -\ln p(\bm{s}|\bm{V}).
\end{equation}

Denote the real video and pseudo video as $\mathbf{V}_r$ and $\mathbf{V}_p$, respectively.
For the two basic streams with real video and pseudo video, we define two CTC loss as alignment loss $\mathcal{L}_A$, written as follows
\begin{equation}\label{equ:loss_align}
  \mathcal{L}_A = \mathcal{L}_{\mathrm{CTC}}^r + \mathcal{L}_{\mathrm{CTC}}^p,
\end{equation}
where $\mathcal{L}_{\mathrm{CTC}}^r$ and $\mathcal{L}_{\mathrm{CTC}}^p$ are the CTC loss for real video and pseudo video, respectively.
The alignment loss targets at maximizing the total probabilities of all alignment paths between the sign video and the label sequence.

In our method, the video data and text label are mapped into the same latent space, which makes it possible for distance measurement.
For the input data modalities, \emph{i.e.,}, real video, real label, pseudo video, pseudo label, the corresponding feature representations are denoted as $\bm{f}_v^r$, $\bm{f}_l^r$, $\bm{f}_v^p$, $\bm{f}_l^p$ from Equation~\eqref{equ:blstm} and Equation~\eqref{equ:text encoder}, respectively.
We divide the features into two groups, which are $(\bm{f}_v^r, \bm{f}_l^r, \bm{f}_l^p)$ and $(\bm{f}_l^r, \bm{f}_v^r, \bm{f}_v^p)$, respectively.
For $(\bm{f}_v^r, \bm{f}_l^r, \bm{f}_l^p)$, the distance of the feature representations between real video and real label is supposed to be closer than that between real video and pseudo label.
That is to say, in such triplet, the feature representation of real video is regarded as the anchor, with the real and pseudo label as a positive and negative sample, respectively.
We use triplet loss as the objective function to minimize the distance from the anchor to the positive sample, and maximize the distance from the anchor to the negative sample.
The real video anchor based real-pseudo discriminative loss is defined as follows, 
\begin{equation}\label{equ:loss_dv}
  \mathcal{L}_{D_v} = \mathcal{L}_{trip}(\bm{f}_v^r, \bm{f}_l^r, \bm{f}_l^p) = \max \left (\mathcal{D}(\bm{f}_v^r, \bm{f}_l^r) - \mathcal{D}(\bm{f}_v^r, \bm{f}_l^p)+\alpha, 0 \right),
\end{equation}
where $\mathcal{L}_{trip}(\cdot)$ means triplet loss~\cite{hoffer2015deep}, 
$\mathcal{D}$ is the distance function, $\alpha$ is a margin.
For another group, with the same purpose, the real text anchor based real-pseudo discriminative loss is defined as follows, 
\begin{equation}\label{equ:loss_dl}
  \mathcal{L}_{D_l} = \mathcal{L}_{trip}(\bm{f}_l^r, \bm{f}_v^r, \bm{f}_v^p) = \max \left (\mathcal{D}(\bm{f}_l^r, \bm{f}_v^r) - \mathcal{D}(\bm{f}_l^r, \bm{f}_v^p) +\alpha, 0 \right).
\end{equation}
The final real-pseudo discriminative loss $\mathcal{L}_{D}$ is the summation of these two parts, written as follows,
\begin{equation}\label{equ:loss_d}
  \mathcal{L}_{D} = \mathcal{L}_{D_v} + \mathcal{L}_{D_l}.
\end{equation}

The real-pseudo discriminative loss focuses on the relative distance between the video and text label data.
For real video-text pair, the distance between the features of such pair in latent space is expected to get as close as possible.
Hence, we directly minimize the distance of the real video-text pair, called cross modality semantic correspondence loss.
The loss function is defined as the distance metric,
\begin{equation}\label{equ:loss_s}
  \mathcal{L}_{S} = \mathcal{D}(\bm{f}_v^r, \bm{f}_l^r),
\end{equation}
where $\mathcal{D}$ is the distance metric function. Considering the lengths of video and text are variable, 
to calculate the distance between two variable length sequences, the distance metric function between $\bm{f}_v^r$ and $\bm{f}_l^r$ is defined as the dynamic time warping (DTW) discrepancy.
Denoting the cost between $\bm{f}_v^r$ and $\bm{f}_l^r$ at different time steps $t_1$ and $t_2$ as $d(t_1,t_2)$,
DTW typically uses dynamic programming to efficiently find the best alignment that minimizes the overall cost.
Define the DTW distance for subsequences $\bm{f}_v^r(1:i)=(\bm{f}_v^r(1), \bm{f}_v^r(2), \cdots, \bm{f}_v^r(i))$ and $\bm{f}_l^r(1:j)=(\bm{f}_l^r(1), \bm{f}_l^r(2), \cdots, \bm{f}_l^r(j))$ as $D_{i,j}$, which can be written as 
\begin{equation}\label{equ:dtw_pice}
  D_{i,j}=d(i,j) + \min (D_{i-1,j}, D_{i,j-1}, D_{i-1,j-1}).
\end{equation}
In our experiments, $d$ is calculated as the cosine distance, written as follows,
\begin{equation}\label{equ:dtw_d_cosine}
  d(i,j)=1 - \frac{\bm{f}_v^r(i) \cdot \bm{f}_l^r(j)} {||\bm{f}_v^r(i)|| \cdot ||\bm{f}_l^r(j)||}.
\end{equation}

To make DTW distance differentiable, a continuous relaxation of the minimum operator \cite{cuturi2017soft,chang2019d3tw} is introduced with a smoothing parameter $\gamma \geq 0$
\begin{equation}
    \min\nolimits^{\gamma}{(a_1, \cdots, a_n)} :=
     \left\{
        \begin{aligned}
            &\min_{i} a_i &\gamma=0. \\
            &-\gamma\log \sum_{i} e^{-a_i/\gamma} &\gamma \geq 0.
        \end{aligned}
     \right.
\end{equation}
With the formulation of DTW, $\mathcal{D}(\bm{f}_v^r, \bm{f}_l^r)$ is calculated as follows,
\begin{equation}
\mathcal{D}(\bm{f}_v^r, \bm{f}_l^r) = D_{T,N},
\end{equation}
where $T$ is the length of $\bm{f}_v^r$ and $N$ is the length of $\bm{f}_l^r$.
The distance in triplet loss used in Equation~\eqref{equ:loss_dv} and Equation~\eqref{equ:loss_dl} is also calculated by DTW since the lengths of the items in triplet are variable.

The final objective loss function is defined as follows,
\begin{equation}\label{equ:loss_all}
  \mathcal{L} = \lambda \mathcal{L}_A + (1-\lambda) (\mathcal{L}_D + \mathcal{L}_S),
\end{equation}
where $\lambda$ is a hyper-parameter, which indicates the weighted summation of these loss terms.
Since $\mathcal{L}_D$ and $\mathcal{L}_S$ have the same distance metrics, we combine them and perform weighted sum with $\mathcal{L}_A$.

\begin{table*}[ht]
\begin{center}
\caption{Statistical data on RWTH-PHOENIX-Weather multi-signer, signer-independent and CSL datasets.}
\label{tab:stat}
\tabcolsep=6pt
\begin{tabular}{l|ccc|ccc|cc}
\hline
\multirow{3}{*}{Statistics} & \multicolumn{3}{c|}{RWTH-PHOENIX-Weather} & \multicolumn{3}{c|}{RWTH-PHOENIX-Weather} & \multicolumn{2}{c}{CSL} \\ 
                           & \multicolumn{3}{c|}{Multi-Signer} & \multicolumn{3}{c|}{Signer-Independent} &   \\ 
                         &  Train & Dev & Test &  Train & Dev & Test & Train  &  Test \\ \hline\hline
\#signers         & 9       & 9      & 9      & 8       & 1      & 1      & 50      & 50 \\
\#frames          & 799,006 & 75,186 & 89,472 & 612,027 & 16,460 & 26,891 & 963,228 & 66,529 \\
\#duration~(h)    & 8.88    & 0.84   & 0.99   & 6.80    & 0.18   & 0.30   & 10.70   & 0.74 \\
\#vocabulary      & 1,231   & 460    & 496    & 1,081   & 239    & 294    & 178     & 20 \\
\#videos          & 5,672   & 540    & 629    & 4,376   & 111    & 180    & 4,700   & 300 \\ \hline

\end{tabular}
\end{center}
\end{table*}

\section{Experiments}
In this section, we conduct comprehensive experiments to validate the effectiveness of our method.
We first review our benchmark datasets and evaluation metrics.
Then we perform ablation studies on each part of our proposed framework.
Finally, we compare our method with state-of-the-art approaches on two benchmark datasets.

\subsection{Dataset and Evaluation}
We perform our experiments on two benchmark datasets, \emph{i.e.,} RWTH-PHOENIX-Weather multi-signer~\cite{koller2015continuous}, RWTH-PHOENIX-Weather signer-independent~\cite{koller2017re} and CSL~\cite{huang2018video} dataset.
RWTH-PHO
ENIX-WEATHER multi-signer dataset, focusing on the German sign language, is one of the most popular benchmark datasets in continuous SLR.
This dataset is recorded from a public television broadcast from a monocular RGB camera at 25 frames per second~(fps), with a resolution of $210\times260$.
It contains a total of 6,841 sentences with a total vocabulary size of 1,295 sign words performed by 9 different signers.
Besides the training set, two independent sets are proposed for evaluation, \emph{i.e.,} dev and test set.
Each set accounts for about 10\% of the size of the training set.
All 9 signers appear in these 3 sets.
Additionally, there is another signer-independent setting,
where it chooses 8 signers for training and leaves out signer \#5 for evaluation.
CSL~\cite{huang2018video} is a continuous Chinese sign language dataset containing 5,000 videos performed by 50 signers.
It contains a total of 100 different sentences with a vocabulary size of 100 sign words in daily life.
This dataset is divided into the training and test set, containing 4,700 and 300 videos, respectively.
The detailed statistics are listed in Table~\ref{tab:stat}.

We utilize multiple evaluation metrics for continuous SLR.
Word error rate~(WER) is one of the commonly used metrics.
It is actually an edit distance, indicating the minimum number of operations, \emph{i.e.,} substitution, deletion and insertion, required to convert the predicted sentence to the reference one:
\begin{equation}\label{equ:wer}
  WER = \frac{n_i + n_d + n_s}{L},
\end{equation}
where $n_i$, $n_d$, and $n_s$ are the number of operations for insertion, deletion, and substitution, respectively.
We also calculate the ratio of correct words to the reference words, denoted as Acc-w.
Besides, we adopt some semantic metrics in Neural Language Processing~(NLP) and Neural Machine Translation~(NMT), including BLEU, METEOR, CIDEr and ROUGE-L.

\subsection{Implementation Details}
In our experiment, we optimize our framework in a staged strategy following the previous methods~\cite{pu2019iterative, cui2019deep}.
First, the backbone GoogLeNet adopts the parameters pre-trained on ILSVRC-2014~\cite{russakovsky2015imagenet} dataset.
For the end-to-end training stage, the whole framework is supervised by the loss in Equation~\eqref{equ:loss_all}.
In the sequential model and text encoder, BLSTMs both have two layers with the hidden states set to 1024.
Adam optimizer is utilized with the learning rate as 5e-3 and batch size as 3.
The network is trained for 50 epochs till convergence.
In continuous SLR, it is crucial for the visual encoder to produce robust feature representation.
The loss terms utilized for the first stage have limited contribution to the low layers of the visual encoder due to the vanishing gradient, which makes the visual encoder not fully optimized.
Therefore, we use the CTC decoding method to generate pseudo labels for video clips.

After that, in the second stage, we utilize these clip-label pairs for classification.
Specifically, we add a fully-connected layer on top of the visual encoder and use the cross-entropy loss to supervise its learning.
It is optimized by stochastic gradient descent~(SGD) with totally 40 epochs.
The initial learning rate is set as 5e-3 and with 10x reduction when loss saturates.
The input clip length is 16.
We set the batch size as 32 and weight decay as 1e-4, respectively.
Embedded with the optimized visual encoder at this stage, we then train the whole framework using the loss in Equation~\eqref{equ:loss_all} again.
Through such optimization strategy, our visual encoder cooperates with the sequential model for better performance.
Our whole framework is implemented on PyTorch and experiments are performed on NVIDIA Tesla V100.

Besides, data augmentation is crucial for relieving over-fitting.
During training, the video is randomly cropped at the same spatial location along the time dimension, with the resolution of $224\times224$.
Then it is randomly flipped horizontally.
Temporally, we randomly discard 20\% frames individually.
During testing, the video is center cropped at the same spatial location with the resolution of $224\times224$.
All the frames in the video are fed into the framework.

\subsection{Ablation Study}
We perform ablation studies on the effectiveness of different parts in our framework.

\begin{figure}[ht]
  \begin{center}
  \includegraphics[width=1\linewidth]{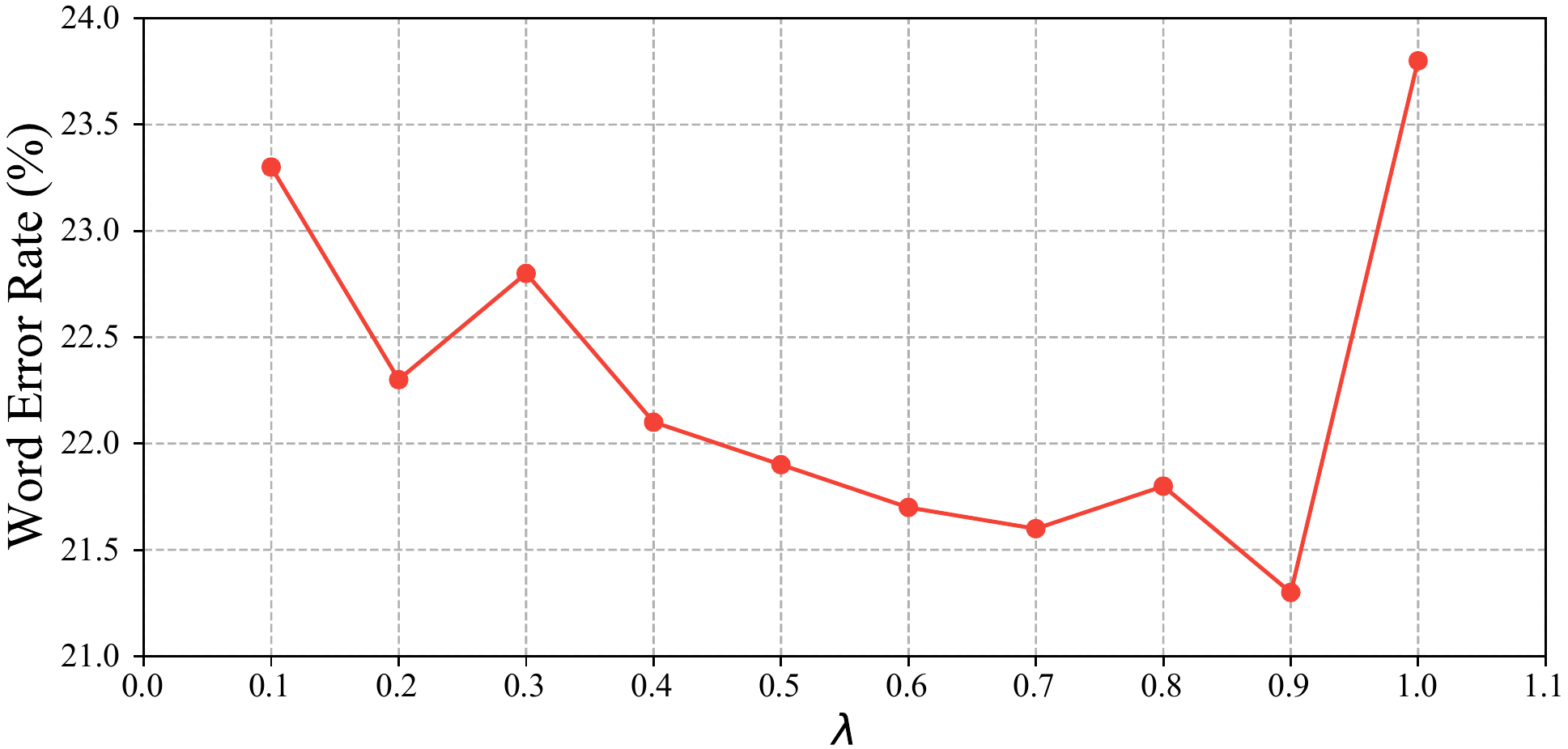}
  \end{center}
  \caption{Effects of different hyper parameter $\lambda$ in Equation~\eqref{equ:loss_all} on the RWTH-PHOENIX-Weather multi-signer dataset.}
  \label{fig:ratio}
  \Description{}
  \vspace{-2pt}
\end{figure}

\textbf{Hyper parameter $\lambda$.}
We study the impact of $\lambda$ in Equation~\eqref{equ:loss_all} and the result is shown in Figure~\ref{fig:ratio}.
The experiments are conducted on RWTH-PHOENIX-Weather multi-signer independent dataset and we utilize the WER result on the dev set as our choosing criterion.
It can be seen that the WER result gets improved with a gradual and flexuous process and achieves the best when $\lambda=0.9$.
Notice that $\lambda=1$ corresponds to baseline which optimized with only CTC loss.
When $\lambda=0$, the network does not converge, so there is no result shown in Figure~\ref{fig:ratio}.
Unless stated, we utilize it as our default hyper parameters in the following experiments.

\begin{figure*}
\centering
\includegraphics[width=1.0\textwidth]{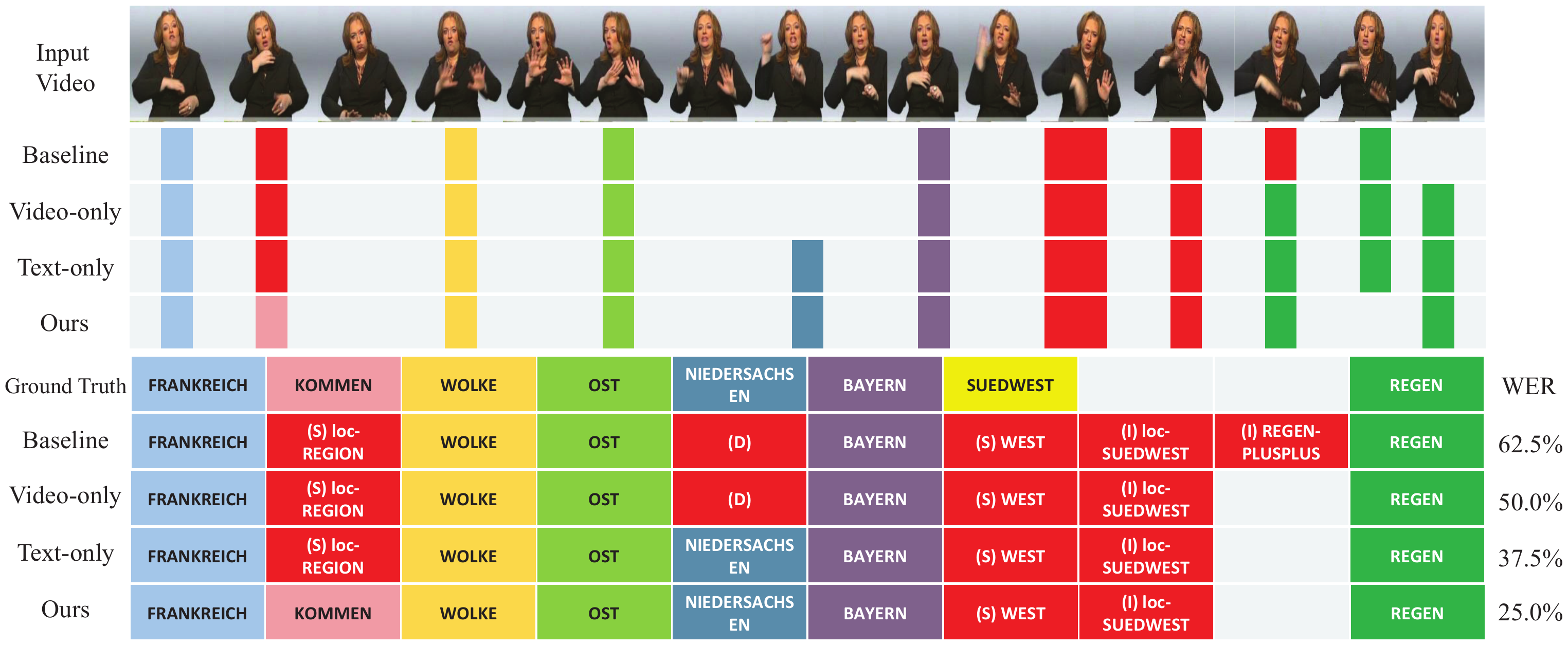}
\caption{An example on the the dev set of RWTH-PHOENIX-Weather multi-signer dataset. In the figure, the first row represents the input frame sequences. The medium part indicates the sign word with maximum probability at each time step. The bottom part shows the final predicted sentence. Red symbols denotes the wrongly predicted words.
``D'', ``S'', and ``I'' stand for deletion, substitution, and insertion, respectively.}
 \label{fig:sample-rwth}
 \Description{}
\end{figure*}

\textbf{Effectiveness on the pseudo video-text pairs.} 
We compare the effectiveness of our pseudo video and text respectively on the RWTH-PHOENIX-Weather multi-signer dataset in Table~\ref{tab:pesu_vtp}.
``Video only'' denotes that we only generate pseudo video after editing and supervise the whole framework by the loss terms except the pseudo text related ones.
It can be observed that the WER result is improved by 1.8\% and 1.9\% over the baseline on the dev and test set, respectively.
``Text only'' denotes that we only generate pseudo text after editing.
The improvement on the WER result is similar to the former, with 1.9\% and 2.0\% on the dev and test set, respectively.
When the pseudo video-text pair is generated after editing, the performance is further improved to 21.3\% and 21.9\% on the dev and test set, respectively.
It can be concluded that the generated video and text are both beneficial for the performance boost.

To qualitatively show its effectiveness, we further visualize an example as shown in Figure~\ref{fig:sample-rwth}.
It can be seen the hypothesis sentence of the baseline method shows deletion, substitution and insertion compared with the ground truth sentence.
With only pseudo video or text inserted into our framework, the WER result gets improved to some extent, \emph{e.g.} ``Video-only'' method corrects the failure insertion of word ``REGEN-PLUSPLUS''.
When using both pseudo video and text, the WER result is further improved by a large margin on this sentence, which also shows the complementary effect of the pseudo video-text pair.

\begin{table}
\begin{center}
\caption{An ablation study on the effectiveness of the pseudo video-text pair on the RWTH-PHOENIX-Weather multi-signer dataset (the lower the better).}
\label{tab:pesu_vtp}
\tabcolsep=8pt
\begin{tabular}{c|rc|rc}
\hline
\multirow{2}{*}{Methods} & \multicolumn{2}{c|}{Dev} & \multicolumn{2}{c}{Test} \\ 
           & del / ins & WER   & del / ins & WER    \\ \hline\hline
Baseline   & 7.8 / 3.5 & 23.8  & 7.8 / 3.4 & 24.4   \\
Video only & 7.7 / 3.0 & 22.0  & 7.0 / 2.8 & 22.5   \\
Text only  & 8.1 / 2.8 & 21.9  & 7.8 / 2.4 & 22.4   \\
Ours       & 7.3 / 2.7 & \textbf{21.3}  & 7.3 / 2.4 & \textbf{21.9}   \\ \hline
\end{tabular}
\end{center}
\vspace{-0.3cm}
\end{table}

\begin{table}
\begin{center}
\caption{Effects of the maximum editing operations on the RWTH-PHOENIX-Weather signer-independent dataset (the lower the better).}
\label{tab:max-num}
\tabcolsep=8pt
\begin{tabular}{c|cccccc}
\hline
$K$ & 1    & 2    & 3    & 4    & 5   & 6 \\ \hline \hline
del     & 7.7  & 7.1  & 7.3  & 8.0  & 8.0  & 7.9 \\
ins     & 2.9  & 3.1  & 2.7  & 2.6  & 2.6  & 2.7 \\
WER     & 21.9 & 21.7 & 21.3 & 21.8 & 21.8 & 21.7 \\ \hline
\end{tabular}
\end{center}
\vspace{-0.3cm}
\end{table}

\textbf{Effects of the maximum editing operations.}
We perform experiments on the effects of the maximum editing operations.
As shown in Table~\ref{tab:max-num}, ``$K$'' indicates the maximum number of editing operations.
The number of editing operations on the original video-text pair is randomly selected in the range from 1 to $K$.
It can be seen that it reaches the lowest WER when the max number of operations is 3.
It can be explained that the lower number of operations makes the framework more concentrate on distinguishing the fine-grained differences between the real and pseudo video-text pairs.
Unless stated, we set the default maximum operations as $K=3$ in the following experiments.

\begin{table*}
\begin{center}
\caption{Evaluation on CSL dataset. ($\uparrow$ indicates the higher the better, while $\downarrow$ indicates the lower the better.)}
\label{tab:csl}
\tabcolsep=6pt
\begin{tabular}{l|cccccccc}
\hline
Methods                         &  Acc-w $\uparrow$  & BLEU-1 $\uparrow$  & CIDEr $\uparrow$ & ROUGE-L $\uparrow$  & METEOR $\uparrow$  & WER $\downarrow$ \\ \hline\hline
ELM~\cite{chen2014using} & 0.175 & 0.376 & 0.028 & 0.120 & 0.388 & 0.987 \\ 
LSTM \& CTC~\cite{graves2006connectionist,hochreiter1997long}  &  0.332  & 0.343  & 0.241  & 0.362    & 0.111   & 0.757 \\ 
S2VT (3-layer)~\cite{venugopalan2015sequence}    &  0.461  & 0.475  & 0.477  & 0.465    & 0.186   & 0.652 \\ 
HLSTM-attn~\cite{guo2018hierarchical}     &  0.506  & 0.508   & 0.605  & 0.503    & 0.205   & 0.641 \\ 
HRF-Fusion~\cite{guo2019hierarchical} & 0.445 & 0.450  & 0.398 & 0.449 & 0.171 & 0.672 \\ 
IAN~\cite{pu2019iterative}     &  0.670  & 0.724    & \textbf{3.946}  & 0.716    & 0.383   & 0.327 \\ \hline \hline
\textbf{Ours} &  \textbf{0.747}  & \textbf{0.784} & 3.006 & \textbf{0.782} & \textbf{0.390} & \textbf{0.245} \\ \hline
\end{tabular}
\end{center}
\end{table*}

\begin{table}
\begin{center}
\caption{Evaluation on RWTH-PHOENIX-Weather multi-signer dataset (the lower the better).}
\label{tab:phoenix}
\tabcolsep=5pt
\begin{tabular}{l|rc|rc}
\hline
\multirow{2}{*}{Methods} & \multicolumn{2}{c|}{Dev} & \multicolumn{2}{c}{Test} \\ 
                         &  del / ins   & WER       & del / ins      & WER      \\ \hline\hline
CMLLR~\cite{koller2015continuous}  & 21.8 / 3.9 & 55.0  & 20.3 / 4.5      & 53.0  \\
1-Million-Hand~\cite{koller2016deep-cvpr}  & 16.3 / 4.6 & 47.1  & 15.2 / 4.6      & 45.1    \\
CNN-Hybrid~\cite{koller2016deep} & 12.6 / 5.1 & 38.3  & 11.1 / 5.7      & 38.8    \\
SubUNets~\cite{camgoz2017subunets} & 14.6 / 4.0 & 40.8  & 14.3 / 4.0      & 40.7    \\
RCNN~\cite{cui2017recurrent}  & 13.7 / 7.3 & 39.4 & 12.2 / 7.5 & 38.7 \\
Re-sign~\cite{koller2017re}   & -  & 27.1  & -     & 26.8    \\
Hybrid CNN-HMM~\cite{koller2018deep} & - &31.6 & - & 32.5 \\ 
CNN-LSTM-HMM~\cite{koller2019weakly}   & - & 26.0  & -  & 26.0    \\
CTF~\cite{wang2018connectionist}  & 12.8 / 5.2 & 37.9  & 11.9 / 5.6  & 37.8    \\
Dilated~\cite{pu2018dilated}   & 8.3 / 4.8 & 38.0  & 7.6 / 4.8  & 37.3    \\
IAN~\cite{pu2019iterative}  & 12.9 / 2.6 & 37.1 & 13.0 / 2.5 & 36.7 \\
DNF (RGB)~\cite{cui2019deep}   & 7.8 / 3.5 & 23.8  & 7.8 / 3.4 & 24.4  \\ \hline \hline
Ours     & 7.3 / 2.7 & \textbf{21.3}  & 7.3 / 2.4 & \textbf{21.9}    \\ \hline
\end{tabular}
\end{center}
\end{table}

\begin{table}
\begin{center}
\caption{Evaluation on RWTH-PHOENIX-Weather signer-independent dataset (the lower the better).}
\label{tab:s-ind}
\tabcolsep=8pt
\begin{tabular}{c|rc|rc}
\hline
\multirow{2}{*}{Methods} & \multicolumn{2}{c|}{Dev} & \multicolumn{2}{c}{Test} \\ 
           & del / ins & WER   & del / ins & WER    \\ \hline\hline
Re-sign~\cite{koller2017re}   & - & 45.1  & - & 44.1   \\
Baseline~\cite{cui2019deep}   &  9.2  / 4.3 & 36.0 & 9.5 / 4.6  & 35.7   \\
Ours       &  11.1 / 2.4 & \textbf{34.8} & 11.4 / 3.3 & \textbf{34.3} \\ \hline
\end{tabular}
\end{center}
\vspace{-0.3cm}
\end{table}

\subsection{Comparison with the State-of-the-art Methods}
We perform extensive experiments and compare with other state-of-the-art methods on two benchmark datasets, including RWTH-PHOENIX-Weather multi-signer, signer-independent and CSL dataset.

\textbf{Evaluation on the RWTH-PHOENIX-Weather multi-sign
er dataset.}
The results on RWTH-PHOENIX-Weather multi-signer dataset are shown in Table~\ref{tab:phoenix}.
CMLLR \cite{koller2015continuous} and 1-Million-Hand \cite{koller2016deep-cvpr} are classical methods using hand-crafted features with traditional HMM models.
CMLLR designs specific features to describe signs from different aspects of SLR, \emph{e.g.} HOG-3D features, trajectories with position, high-level face features and temporal derivatives.
With the advance of deep learning, researchers utilize CNNs to adaptively extract feature representations with significant performance gain.
SubUNets \cite{camgoz2017subunets} solves simultaneous alignment and recognition problems in an end-to-end framework by incorporating the CNN-BLSTM architecture supervised by the CTC loss.
Re-sign \cite{koller2017re} presents an iterative re-alignment approach with further embedding a HMM to correct the frame labels and continuously improves its performance.
DNF \cite{cui2019deep} explores the suitable CNN-BLSTM framework and the function of multiple input modalities, becoming the most competitive method.
Even compared with these challenging methods, our method still achieves a new state-of-the-art result in this dataset, \emph{i.e.,} 21.3\% and 21.9\% WER on the dev and test set, respectively. 
It surpasses the best competitor with 2.6\% and 2.5\% on the dev and test set, respectively.

\textbf{Evaluation on the RWTH-PHOENIX-Weather signer-
independent dataset.}
The signer-independent subset is created based on the RWTH-PHOENIX-Weather, where we test on a single individual who has not been seen during training.
In Table~\ref{tab:s-ind}, we compare existing the methods on this dataset.
It can be observed that our method still outperforms all the methods in this dataset, achieving 34.8\% and 34.3\% WER on the dev and test set, respectively.
Compared with the best WER results on its multi-signer counterpart, the independent setting is a more challenging one, with over 10\% WER reduction.

\textbf{Evaluation on the CSL dataset.}
We perform experiments on the challenging split of the CSL dataset in Table~\ref{tab:csl}.
This split is difficult due to the unseen combination and occurrence of words, and different semantic context in the test set.
We compare our approach with other challenging methods, such as HRF-Fusion~\cite{guo2019hierarchical}, HLSTM-atten~\cite{guo2018hierarchical} and IAN \cite{pu2019iterative}.
HLSTM-atten treats this task as sign language translation and proposes a hierarchical-LSTM (HLSTM) encoder-decoder model with visual content and word embedding, and utilize temporal attention for performance boost.
IAN proposes to use the visual encoder and encoder-decoder sequence learning network with iterative refinement.
Compared with these methods, our method also achieves the state-of-the-art performance on most of the evaluation metrics on this dataset.
It should be noted that our method mimics the editing process and surpasses the best competitor with 8.2\% on the WER result, which is consistent with our optimization target.
Besides, it also achieves the state-of-the-art performance on most semantic evaluation metrics, such as BLEU-1, ROUGE-L, METEOR, \emph{etc.}

\section{Conclusion}
In this paper, we attempt to tackle the issue of inconformity between the CTC objective and the evaluation metric via cross modality augmentation.
Following the operations in the definition of WER, \emph{i.e.,} substitution, deletion and insertion, we edit the real video-text pair to generate its corresponding pseudo counterpart.
Besides constraining the semantic correspondence between the video and text, we design a discriminative loss to make the network aware of the differences between the real and pseudo video-text pair.
Our proposed framework can be easily extended to other existing CTC based continuous SLR networks.
We conduct experiments on two continuous SLR benchmarks, \emph{i.e.,} RWTH-PHOENIX-Weather and CSL dataset.
Experimental results validate the effectiveness of our proposed method with notable performance gain over previous methods, especially on the WER metric.

\begin{acks}
This work was supported in part to Dr. Houqiang Li by NSFC under contract No. 61836011, and in part to Dr. Wengang Zhou by NSFC under contract No. 61822208 \& 61632019 and Youth Innovation Promotion Association CAS (No. 2018497).
\end{acks}

\bibliographystyle{ACM-Reference-Format}
\bibliography{ref}

\appendix

\end{document}